\documentclass{article}
\usepackage{amsmath} 
\usepackage{graphicx} 
\usepackage{caption}
\usepackage{subfigure}

\PassOptionsToPackage{numbers, compress}{natbib}
\usepackage[final]{tackling_climate_workshop_style}

\usepackage[utf8]{inputenc} % allow utf-8 input
\usepackage[T1]{fontenc}    % use 8-bit T1 fonts
\usepackage{hyperref}       % hyperlinks
\usepackage{url}            % simple URL typesetting
\usepackage{booktabs}       % professional-quality tables
\usepackage{amsfonts}       % blackboard math symbols
\usepackage{nicefrac}       % compact symbols for 1/2, etc.
\usepackage{microtype}      % microtypography
\graphicspath{ {figures/} }
\title{
CuMoLoS-MAE: A Masked Autoencoder for\\Remote Sensing Data Reconstruction
}

\author{%
\parbox{\textwidth}{
\centering
Anurup Naskar$^{1}$ \quad Nathanael Zhixin Wong$^{1}$ \quad Sara Shamekh$^{1}$ } \vspace{0.2cm} \\ 
$^1$ Courant Institute of Mathematical Sciences, New York University\\
\texttt{\{an4462,nw2648,ss18284\}@nyu.edu}}

\begin{document}

\maketitle

\begin{abstract}
Accurate atmospheric profiles from remote sensing instruments such as Doppler Lidar, Radar, and radiometers are frequently corrupted by low‐SNR (Signal to Noise Ratio) gates, range folding, and spurious discontinuities. Traditional gap filling blurs fine-scale structures, whereas deep models lack confidence estimates. We present \textbf{CuMoLoS-MAE}, a \underline{Cu}rriculum-Guided \underline{Mo}nte Car\underline{lo} \underline{S}tochastic Ensemble \underline{M}asked \underline{A}uto\underline{e}ncoder designed to (i) restore fine-scale features such as updraft and downdraft cores, shear lines, and small vortices, (ii) learn a data-driven prior over atmospheric fields, and (iii) quantify pixel-wise uncertainty. During training, CuMoLoS-MAE employs a mask–ratio curriculum that forces a ViT decoder to reconstruct from progressively sparser context. At inference, we approximate the posterior predictive by Monte Carlo over random mask realisations, evaluating the MAE multiple times and aggregating the outputs to obtain the posterior predictive mean reconstruction \(\bar{X}\) together with a finely resolved per-pixel uncertainty map \(\sigma_X\). Together with high-fidelity reconstruction, this novel deep learning-based workflow enables enhanced convection diagnostics, supports real-time data assimilation, and improves long-term climate reanalysis. 
% Code accessible at \url{https://github.com/anurup123/CuMoLoS-MAE}.

\end{abstract}
\enlargethispage{2\baselineskip}

\section{Introduction}

Understanding climate change's impacts on extreme events requires accurate and continuous measurements from remote sensing instruments such as Doppler-lidar, radar, or radiometers. However these data often suffer from missing or corrupted returns that must be filled in before they can be reliably used in downstream applications. Classical gap‐filling procedures, such as sliding‐window mean filters~\cite{lesti2017sliding}, smear out critical small‐scale features like shear lines and updraft cores. Recent advances in deep‐learning approaches (e.g.\ variational autoencoders~\cite{pinheiro2021variational}) can recover sharper structures but provide no information to gauge reconstruction uncertainties when these methods are used in downstream assimilation and alerting frameworks.

To tackle these shortcomings, we propose \textbf{CuMoLoS-MAE}, a \underline{Cu}rriculum-Guided \underline{Mo}nte Car\underline{lo} \underline{S}tochastic Ensemble Masked Autoencoder, a novel approach that comprises three core mechanisms: (1) curriculum masking~\cite{madan2024cl} to stabilise training and encourage reconstruction from sparser context, (2) micro-patch based on masked autoencoder~\cite{he2022masked} to capture fine structure and mid-scale dynamics, and (3) Monte-Carlo~\cite{harrison2010introduction} ensembles to produce per-pixel uncertainty maps alongside the final reconstructions. In this paper, we have used Doppler-lidar measurements from the Southern Great Plains (SGP) provided by the Atmospheric Radiation Measurement (ARM)~\cite{sisterson2016arm} as a case study, and present our reconstructions of the vertical velocity field and their corresponding uncertainty quantification.

\clearpage
\section{Data}

% We use vertical velocity measurements from a Doppler lidar deployed at the ARM Southern Great Plains (SGP) site. The data is stored in NetCDF format as two-dimensional arrays, with time along one axis and vertical range along the other. Each file contains 320 vertical range gates, spaced evenly at 30~m intervals. To focus on the region closest to the surface where convective processes are most active, we extract only the lowest 64 range gates, corresponding to the first 1.92~km of the atmosphere. From this subset, we construct input samples by dividing the time–height arrays into non-overlapping $64\times64$ patches. These patches serve as the fundamental units for training and evaluation in our reconstruction and uncertainty estimation pipeline. We have used data from June~1–9, 2011 for training, and evaluate model performance on an unseen day: June~15, 2011.
We have used vertical velocity data from a Doppler lidar at the ARM SGP site, stored as time–height arrays in NetCDF format with 320 range gates at 30~m spacing, where each range gate represents the mean return from a fixed vertical layer. Focusing on the lowest 64 gates (1.92~km), we extract non-overlapping $64\times64$ patches as inputs for training and evaluation. The model is trained on data from June~1--9, 2011 and evaluated on an unseen day: June~15, 2011.

\section{Methodology}

\subsection{CuMoLoS\textendash MAE Architecture}
\begin{figure}[t]
  \centering
  \includegraphics[
    width= \linewidth
  ]{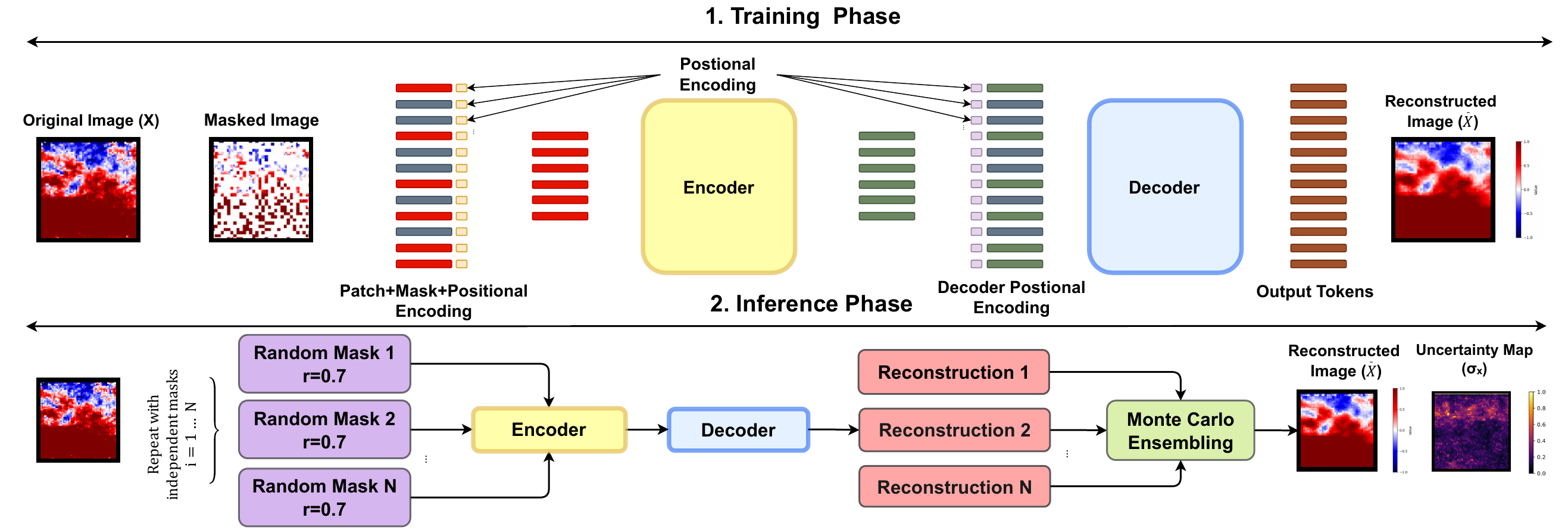}
 \captionsetup{font=footnotesize} 
 \caption{From Doppler lidar time–height arrays we form $64\times64$ images. During training we randomly hide a subset of patch tokens (the mask ratio increases from 0.5 to 0.7 over a curriculum), add positional encodings, pass the visible tokens through a ViT encoder, and reconstruct the full field with a lightweight decoder. The loss is computed only on the hidden pixels. At test time, for each unseen patch we draw 50 independent random masks and run the same pipeline to produce multiple reconstructions. We then average these reconstructions to obtain a single best denoised estimate of the field and use the pixel-wise spread of the ensemble as an uncertainty map.}
  \label{fig:smcmae-example}
\end{figure}
Inspired by masked autoencoding and stochastic ensembling methods for uncertainty quantification, the CuMoLoS-MAE framework is comprised of two stages as shown in Figure~\ref{fig:smcmae-example}:

% \begin{enumerate}
%   \item \textbf{Micro\textendash patchified MAE with curriculum:} Each \(64 \times 64\) Doppler\textendash velocity slice is tokenized into \(2 \times 2\) micro\textendash patches. A 12\textendash layer ViT~\cite{han2022survey} encoder processes visible tokens and a 4\textendash layer decoder reconstructs the field. During training, the mask ratio \(\rho\) starts at \(50\%\) for 5 epochs, then cosine\textendash ramps to \(70\%\) by epoch 30 and is held thereafter; optimization uses masked\textendash MSE on the hidden pixels.

%   \item \textbf{Monte Carlo ensembling:} At inference time, a new random mask is generated for each forward pass and repeat the masking, encoding, and decoding process \(N = 50\) times. The resulting ensemble of predictions is then aggregated to compute the mean and pixel-wise standard deviation:
% \begin{align}
%   \bar X &= \frac{1}{N}\sum_{i=1}^N \hat X^{(i)}, \quad
%   \sigma_X = \sqrt{\frac{1}{N}\sum_{i=1}^N\bigl(\hat X^{(i)} - \bar X\bigr)^2}, \label{eq:mc}
% \end{align}
% yielding a high-fidelity reconstruction \(\bar X\) and calibrated, per-pixel uncertainty estimate \(\sigma_X\).
% \end{enumerate}

\noindent
$\bullet$ \textbf{Micro–patchified MAE with curriculum:} Each \(64 \times 64\) Doppler–velocity slice is tokenized into \(2 \times 2\) micro–patches. A 12–layer ViT~\cite{han2022survey} encoder processes visible tokens and a 4–layer decoder reconstructs the field. During training, the mask ratio $r$ starts at \(50\%\) for 5 epochs, then cosine–ramps to \(70\%\) by epoch 30 and is held thereafter; optimization uses masked–MSE on the hidden pixels.

\vspace{0.5em}
\noindent
$\bullet$ \textbf{Monte Carlo ensembling:} At inference time, a new random mask is generated for each forward pass and the masking, encoding, and decoding process is repeated \(N = 50\) times. The resulting ensemble of predictions is then aggregated to compute the mean and pixel-wise standard deviation:
\begin{align}
  \bar X &= \frac{1}{N}\sum_{i=1}^N \hat X^{(i)}, \quad
  \sigma_X = \sqrt{\frac{1}{N}\sum_{i=1}^N\bigl(\hat X^{(i)} - \bar X\bigr)^2} \label{eq:mc}
\end{align}
yielding a high-fidelity denoised reconstruction \(\bar X\) and per-pixel uncertainty estimate \(\sigma_X\).

\subsection{Training Procedure}

% We preprocess the data by applying an SNR filter (intensity $\geq 0.005$) and clamping valid velocities to the range $[-5, 5]\,\mathrm{m\,s^{-1}}$. From the resulting fields, we extract non\mbox{-}overlapping $64\times64$ patches. Training proceeds over 500 epochs on a single NVIDIA A100 GPU using the AdamW optimiser~\cite{llugsi2021comparison} (base learning rate $1.5\times10^{-4}\cdot\tfrac{32}{256}$, weight decay 0.05) with a batch size of 32. We employ a cosine learning rate schedule with warmup aligned to the mask-ratio curriculum period.\\
The data are preprocessed by applying an SNR filter (intensity $\geq 0.005$) and clamping valid velocities to the range $[-5, 5]\,\mathrm{m\,s^{-1}}$. From the resulting fields, non\mbox{-}overlapping $64\times64$ patches are extracted. Training is conducted over 500 epochs on a single NVIDIA A100 GPU using the AdamW optimiser~\cite{llugsi2021comparison} (base learning rate $1.5\times10^{-4}\cdot\tfrac{32}{256}$, weight decay 0.05) with a batch size of 32. A cosine learning rate schedule with warmup aligned to the mask-ratio curriculum period is employed.

% We preprocess each raw volume by applying an SNR threshold (intensity\(\ge0.005\)), then extract non‐overlapping \(64\times64\) patches (with values spanning \([-5,5]\)m/s). During training, we employ a mask–ratio curriculum that linearly increases the fraction of masked micro‐patch tokens from 0\% to 70\% over the first 30 epochs (holding at 70\% thereafter). We train for 400 epochs on a single NVIDIA A100 GPU using AdamW (base learning rate \(1.5\times10^{-4}\times\frac{32}{256}\), weight decay 0.05) with cosine warmup across the curriculum period, and a batch size of 32.  

 % a \(3\times3\) jump filter (\(\lvert\Delta v\rvert>8\,\mathrm{m/s}\))

% \subsection{Uncertainty map}
% \subsection{Uncertainty map}
% At inference, we draw \(N=50\) independent random masks per image, pass each through the decoder, and aggregate the ensemble (Eq.~\eqref{eq:mc}) to obtain the posterior mean \(\bar X\) and the pixel-wise standard deviation \(\sigma_X\). The value of N is chosen in order to balance the stability of the ensemble with the computational cost. 
\subsection{Uncertainty map}
At inference, \(N=50\) independent random masks are drawn per image, each masked input is passed through the decoder, and the resulting ensemble is aggregated (Eq.~\eqref{eq:mc}) to obtain the posterior mean \(\bar X\) and the pixel-wise standard deviation \(\sigma_X\). The value of N is chosen to balance ensemble stability with computational cost.

% This uncertainty map can be used to identify low-quality regions (e.g., low-SNR or jump-filtered gates), select reliable pixels via risk–coverage analysis, apply confidence-based weighting in data assimilation, or perform adaptive smoothing while preserving storm-resolving features.
% (\(R = \mathrm{diag}(\sigma_X^2 + \varepsilon)\))

% \paragraph{Workflow overview (Fig.~\ref{fig:smcmae-example}).}
% From Doppler lidar time–height arrays we form $64\times64$ images. During training we randomly hide a subset of patch tokens (the mask ratio increases from 0.5 to 0.7 over a curriculum), add positional encodings, pass the visible tokens through a ViT encoder, and reconstruct the full field with a lightweight decoder. The loss is computed only on the hidden pixels. At test time, for each unseen patch we draw 50 independent random masks and run the same pipeline to produce multiple reconstructions. We then average these reconstructions to obtain a single best estimate of the field and use the pixel-wise spread of the ensemble as an uncertainty map, where larger spread indicates lower confidence.

% \paragraph{What the figure shows.}
% Left: an example ground-truth $64\times64$ patch and a masked input. Middle: encoder–decoder flow used in training and reused in inference. Right: the two outputs we deliver for every patch:
% (i) the ensemble-mean reconstruction $\bar X$ (our best estimate of the true field) and
% (ii) the per-pixel uncertainty map $\sigma_X$ (higher values indicate lower confidence)

\section{Results}

% On 1028 held-out images, CuMoLoS-MAE outperforms the 8×8 mean filter, Noise2Void U-Net~\cite{song2021noise2void}, DnCNN (Noise2Void) ~\cite{zhao2018low} and a convolutional variational autoencoder (CVAE)~\cite{bao2017cvae} baseline (Table \ref{tab:quantitative}), while achieving the highest low-frequency spectral fidelity of 93.75\%. We assess spectral integrity via

% \[
% \varepsilon_{\log}(f) = \frac{\log_{10}P_{\mathrm{den}}(f) - \log_{10}P_{\mathrm{raw}}(f)}{\log_{10}P_{\mathrm{raw}}(f)}
% \]

% requiring \(\lvert\varepsilon_{\log}(f)\rvert \le 0.5\) for \(f \le 0.01\ \mathrm{Hz}\). Despite lower performance on all other metrics, the DnCNN (Noise2Void)  attains the best FID (0.12), likely due to perceptual similarity in low-variance regions. In our case, 93.75\% of the low-frequency bins satisfy this criterion, indicating preservation of storm-scale energy. Gate 0 is the principal exception, likely owing to its location at the base of the atmospheric boundary layer as shown in Figure~\ref{fig:results-ebm-returns}.  All baselines have used the same search budget and training protocol on identical data splits, with the same loss, and preprocessing.

\noindent
On the 1028 held-out test set, CuMoLoS-MAE surpasses the 8\(\times\)8 mean filter, Noise2Void U-Net~\cite{song2021noise2void}, Noise2Void Denoising CNN (DnCNN)~\cite{zhao2018low}, and a Convolutional Variational Autoencoder (CVAE~\cite{bao2017cvae})(Table~\ref{tab:quantitative}). It also attains the highest low-frequency spectral fidelity (\(93.75\%\)). Spectral integrity compares temporal Power Spectral Density of the denoised output \(P_{\mathrm{den}}(f)\) and raw signal \(P_{\mathrm{raw}}(f)\) via:
\begin{equation}
\varepsilon_{\log}(f)=\frac{\log_{10}P_{\mathrm{den}}(f)-\log_{10}P_{\mathrm{raw}}(f)}{\log_{10}P_{\mathrm{raw}}(f)}.
\end{equation}
Fidelity is the fraction of bins with \(|\varepsilon_{\log}(f)|\le 0.5\) for \(f\le 0.01\,\mathrm{Hz}\). CuMoLoS-MAE satisfies this in \(93.75\%\) of low-frequency bins, preserving storm-scale energy. Gate~0 is the principal exception, likely owing to its location at the base of the atmospheric boundary layer as shown in Figure~\ref{fig:results-ebm-returns}. The DnCNN (Noise2Void) baseline yields the best FID (0.12) despite worse reconstruction and lower spectral fidelity, likely due to perceptual similarity in low-variance regions.

\begin{table}[ht]
  \centering
  \captionsetup{font=footnotesize}
  \caption{Reconstruction performance and low‐frequency spectral fidelity on 1028 held‐out radar patches. Up arrow (↑) indicates higher is better; down arrow (↓) indicates lower is better.}\vspace{0.1cm}
  \begin{tabular}{lccccc}
    \toprule
    \textbf{Method}            & \textbf{PSNR (dB) ↑} & \textbf{SSIM ↑} & \textbf{MSE ↓} & \textbf{FID ↓} & \textbf{Spectral Fidelity ↑} \\
    \midrule
    8×8 Mean‐Filter            & 23.41           & 0.4950          & 0.5186         & 5.13           & 91.67\% \\
    CVAE                        & 26.70           & 0.4190          & 0.4036         & 3.28           & 80.21\% \\
    DnCNN (Noise2Void) & 23.09           & 0.6466          & 0.6232         & \textbf{0.12}  & 36.46\% \\
    U-Net (Noise2Void)        & 27.70           & 0.7016          & 0.2581         & 0.44           & 49.48\% \\
    \textbf{CuMoLoS-MAE (ours)} & \textbf{29.45}  & \textbf{0.7857} & \textbf{0.1854} & 1.87           & \textbf{93.75\%} \\
    \bottomrule
  \end{tabular}
  \label{tab:quantitative}
\end{table}

\subsection{Uncertainty Quantification and Climate Utility}
\begin{figure*}[htbp]
  \centering
  \includegraphics[width=\textwidth]{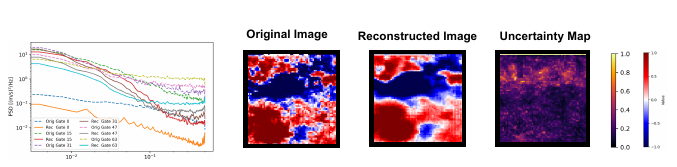}
  \captionsetup{font=footnotesize}
% \caption{Visual diagnostics for a representative sample. Left: power spectral density (log-scaled) for selected range gates. Centre-left: original Doppler-lidar vertical velocity. Centre-right: \textbf{CuMoLoS-MAE} reconstruction. Right: pixel-wise uncertainty map (\(\sigma_X\)).}

\caption{Visual diagnostics for one sample. Left: log-PSD at selected range gates—reconstruction matches the original (small gaps) except at Gate~0 near the boundary-layer base. Centre-left: original Doppler-lidar vertical velocity. Centre-right: \textbf{CuMoLoS-MAE} reconstruction. Right: per-pixel uncertainty map \(\sigma_X\).}

  \label{fig:results-ebm-returns}
\end{figure*}

The reliability of the uncertainty estimates is assessed by computing the pixel-wise Pearson correlation between the Monte Carlo standard deviation map \(\sigma_X\) and the absolute reconstruction error. Across 1028 test images, strong alignment is observed, with a mean per-patch correlation of \(r = 0.961 \pm 0.037\), a global correlation of \(r = 0.961\), and a Spearman rank correlation of \(\rho = 0.926\). As illustrated in Figure~\ref{fig:results-ebm-returns}, these high correlations indicate that \(\sigma_X\) closely tracks the true reconstruction error, i.e., the model reliably predicts where its outputs are likely to be accurate and where they are not. These pixels are split into ten bins by \(\sigma_X\) and the Mean Absolute Error (MAE) in each is computed. The MAE rises monotonically from 0.02845 to 0.99939 across \(\sigma\)-deciles (a \(35.1\times\) gap), and the top 1, 5, 10, and 20 percent of pixels by \(\sigma\) capture 10.1, 30.6, 43.4, and 59.4 percent of total \(\lvert \text{error} \rvert\), confirming the utility of \(\sigma_X\) for error triage in downstream assimilation and warning systems.

\subsection{Effect of Temporal Context on Reconstruction and Spectral Fidelity}

\begin{table}[ht]
  \centering
  \captionsetup{font=footnotesize}
  \caption{Reconstruction quality and spectral fidelity across different temporal window sizes. Up arrow (↑) indicates higher is better; down arrow (↓) indicates lower is better.}\vspace{0.1cm}
  \label{tab:window_comparison}
  \begin{tabular}{lccccc}
    \toprule
    \textbf{Window (time x range gate)} & \textbf{PSNR (dB) ↑} & \textbf{SSIM ↑} & \textbf{MSE ↓} & \textbf{FID ↓} & \textbf{Spectral Fidelity ↑} \\
    \midrule
    $64\times64$   & 29.45  & 0.7857 & 0.1854 & 1.87 & 93.75\% \\
    $128\times64$  & 30.11  & 0.7697 & 0.2253 & 3.73 & 87.50\% \\
    $256\times64$  & 28.55  & 0.6103 & 0.3205 & 5.50 & 38.02\% \\
    \bottomrule
  \end{tabular}
\end{table}

From Table~\ref{tab:window_comparison}, reconstruction quality and low–frequency spectral fidelity decline as the window increases from \(64\times 64\) to \(128\times 64\) and then \(256\times 64\), indicating a predominantly local denoising regime in which \(64\times 64\) provides sufficient context. The slight PSNR bump at \(128\times 64\) arises from a few near-constant, ultra–low-error intervals overweighted by the log average (MSE remains nearly unchanged), which is not statistically significant for any meaningful inference. At \(256\times 64\), token count and masked area rise without added capacity, increasing errors across metrics. 
% Larger windows therefore degrade the reconstruction.

\subsection{Effect of Mask-Ratio Curriculum on Training Efficiency}

% Curriculum masking aims to accelerate convergence while maintaining perceptual quality. We find that the reconstruction loss drops below $0.20$ by epoch $198$ for curriculum masking vs.\ $224$ for fixed masking (${-}26$ epochs; $13$~min), and reaches $0.189$ by epoch $286$ vs.\ $333$ (${-}47$ epochs). The results are comparable according to several different metrices (Table \ref{tab:curriculum_effect}), so we conclude that curriculum masking offers significantly improved training efficiency ($\sim$10\%) over the fixed baseline despite slightly better metrics for the latter methodology.

Curriculum masking accelerates convergence while preserving perceptual quality. Reconstruction loss falls below $0.20$ by epoch $198$ (vs.\ $224$ for fixed; 13~min faster) and reaches $0.189$ by epoch $286$ (vs.\ $333$). As shown in Table~\ref{tab:curriculum_effect}, metrics are comparable, so we conclude that curriculum masking improves training efficiency by $\sim$10\% despite marginally better final scores with fixed masking.

\begin{table}[ht]
\centering
\captionsetup{font=footnotesize}
\caption{Effect of mask-ratio curriculum on reconstruction, FID, and spectral fidelity (1028 patches). Up arrow (↑) indicates higher is better; down arrow (↓) indicates lower is better.}\vspace{0.1cm}
\label{tab:curriculum_effect}
\begin{tabular}{lccccc}
\toprule
\textbf{Configuration}             & \textbf{PSNR (dB)↑} & \textbf{MSE ↓}  & \textbf{SSIM ↑} & \textbf{FID ↓} & \textbf{Spectral Fidelity ↑} \\
\midrule
Without curriculum                 & 29.45              & 0.1854        & 0.7857        & 1.87         & 93.75\%                    \\
With mask-ratio curriculum         & 28.90              & 0.2106        & 0.7868        & 1.88         & 93.23\%                    \\
\bottomrule
\end{tabular}
\end{table}

\section{Conclusion and Future Work}
% \enlargethispage{2\baselineskip}

We have introduced CuMoLoS-MAE, a curriculum-guided masked autoencoder with Monte Carlo ensembling, which delivers state-of-the-art reconstructions together with per-pixel uncertainty. By recovering fine-scale features lost to noise or masking and flagging low-confidence regions, CuMoLoS-MAE improves detection of atmospheric coherent structures. This capability is valuable for Earth observation and climate modelling, as more accurate reconstructed fields improve (1) our understanding of extreme events and their evolution under global warming, (2) data assimilation in numerical weather prediction systems, and (3) process-level analysis within atmospheric sciences. Uncertainty maps further enable reconstitution by weighting observations according to confidence.

Specifically, we plan to use this model to create datasets of vertical profiles to study the shallow-to-deep transition of convection. Reconstructing denoised vertical profiles and their associated uncertainties will help with (a) updraft detection, (b) better characterize their width and kinetic energy, and (c) improve predictors of the transition timing and likelihood. We will evaluate the generalizability of our model across lidar systems provided by ARM (e.g. new datasets from Bankhead National Forest) and assess real-time deployment for operational assimilation. We also plan to scale training from days to months and years to capture climatological variability, support long-term reanalysis, and test robustness under changing environmental conditions.

\clearpage
\bibliographystyle{vancouver}
\bibliography{bibliography}

\end{document}